\documentclass{article}

\usepackage[nonatbib,final]{neurips_2024}

\usepackage[utf8]{inputenc} 
\usepackage[T1]{fontenc}    
\usepackage{hyperref}       
\usepackage{url}            
\usepackage{booktabs}       
\usepackage{amsfonts}       
\usepackage{nicefrac}       
\usepackage{microtype}      
\usepackage{xcolor}         
\usepackage{graphicx}
\usepackage{textcomp}
\usepackage{algorithmicx}
\usepackage{algorithm}
\usepackage{array}
\usepackage{times}  
\usepackage{helvet}  
\usepackage{courier}  
\usepackage{multirow}
\usepackage{graphicx} 
\usepackage{caption} 
\usepackage{colortbl}
\usepackage{algpseudocode}
\usepackage{amsmath}
\usepackage[noabbrev,nameinlink]{cleveref}
\title{Leveraging Hallucinations to Reduce Manual Prompt Dependency in Promptable Segmentation}

\author{
  Jian Hu\textsuperscript{1}, Jiayi Lin\textsuperscript{1}, Junchi Yan\textsuperscript{2}, Shaogang Gong\textsuperscript{1} \\
  \textsuperscript{1}Queen Mary University of London, \textsuperscript{2}Shanghai Jiao Tong University \\
  \{jian.hu, jiayi.lin, s.gong\}@qmul.ac.uk, yanjunchi@sjtu.edu.cn\\
\textcolor{purple!80}{https://lwpyh.github.io/ProMaC/}
}

\begin{document}

\renewcommand{\thefootnote}{\fnsymbol{footnote}}
\footnotetext{This work was supported by Veritone, Adobe, OpenAI,  the Apocrita HPC facility from the QMUL Research-IT, NSFC (92370201, 62222607) and Shanghai Municipal Science and Technology Major Project under Grant 2021SHZDZX0102. Thanks to Weitong Cai for helpful discussion.}
\renewcommand{\thefootnote}{\arabic{footnote}}

\maketitle
\vspace{-10pt}
\begin{abstract}
Promptable segmentation typically requires instance-specific manual prompts to guide the segmentation of each desired object.
To minimize such a need, task-generic promptable segmentation has been introduced, which employs a single task-generic prompt to segment various images of different objects in the same task.
Current methods use Multimodal Large Language Models (MLLMs) to reason detailed instance-specific prompts from a task-generic prompt for improving segmentation accuracy. 
The effectiveness of this segmentation heavily depends on the precision of these derived prompts. 
However, MLLMs often suffer from hallucinations during reasoning, resulting in inaccurate prompting. 
While existing methods focus on eliminating
hallucinations to improve a model, we argue that MLLM hallucinations can reveal valuable contextual insights when leveraged correctly, as they represent pre-trained large-scale knowledge beyond individual images.
In this work, we utilize hallucinations to mine task-related information from images and verify its accuracy for enhancing precision of the generated prompts.  
Specifically, we introduce an iterative \textbf{Pro}mpt-\textbf{Ma}sk \textbf{C}ycle generation framework (ProMaC) with a prompt generator and a mask generator. 
The prompt generator uses a multi-scale chain of thought prompting, initially exploring hallucinations for extracting extended contextual knowledge on a test image. 
These hallucinations are then reduced to formulate precise instance-specific prompts, directing the mask generator to produce masks that are consistent with task semantics by mask semantic alignment.
The generated masks iteratively induce the prompt generator to focus more on task-relevant image areas and reduce irrelevant hallucinations, jointly resulting in better prompts and masks.
Experiments on 5 benchmarks demonstrate the effectiveness of ProMaC. Code given in \textcolor{purple!80}{https://lwpyh.github.io/ProMaC/}.
\end{abstract}
\section{Introduction}
\label{sec:introduction}
Current promptable segmentation methods rely on instance-specific
manual prompts to guide segmentation, greatly limiting its large-scale
application.  
Recently, a manual-free task-generic promptable segmentation approach
was introduced \cite{hu2023relax}: only a single task-generic prompt is
needed for all samples under the same task, e.g.,
``camouflaged animal'' is a task-generic prompt for all images in
a camouflaged object detection task.    
The model segments task-relevant objects in various images based on this generic prompt, significantly reducing the annotation workload.

\begin{figure*}[ht]
  \centering
  \begin{tabular}{cc} 
\hspace{-4mm}\includegraphics[width=0.335\columnwidth]{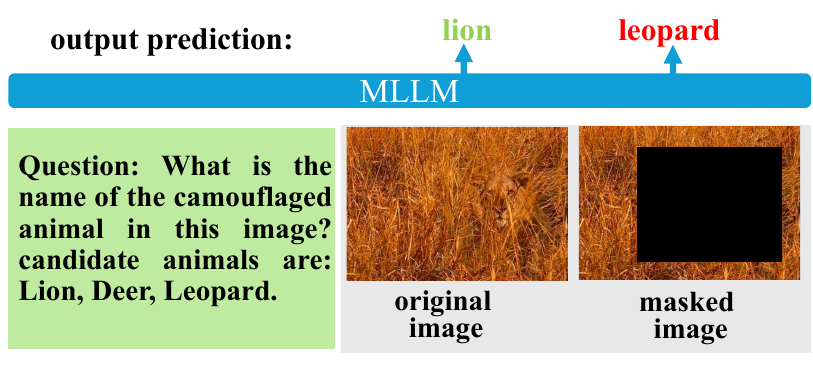} & \hspace{-5mm}\includegraphics[width=0.654\columnwidth]{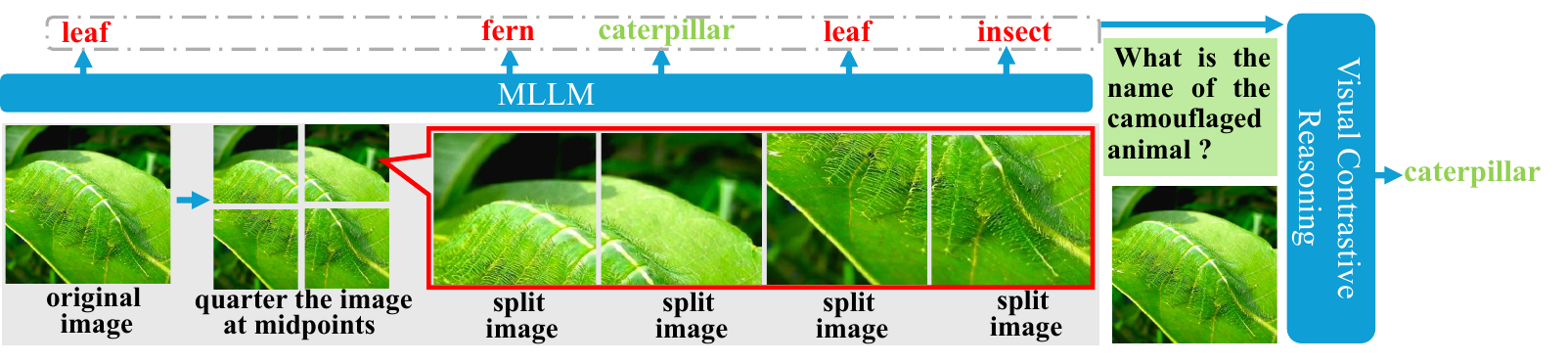}\vspace{-2pt}
\\
    \hspace{-4mm}  
    {\small{\selectfont(a) Hallucination by co-occurrence prior.}} & \hspace{-4mm}  
    {\small{\selectfont(b) Using hallucinations can benefit accuracte prompt generation.}} 

  \end{tabular}
   \caption{
   (a) During MLLM pretraining, leopards often co-occur with grass. If the lion is masked, the model incorrectly identifies it as a leopard based on the grass. 
   %
   (b) Directly inputting the image into MLLM causes the hidden caterpillar being incorrectly predicted as a leaf. Splitting the image results in interested objects being incomplete or absent, prompting MLLM to induce hallucinations and utilize prior knowledge to predict potential task-related objects within the image. Our visual contrastive reasoning then eliminates the hallucinations and validates the gathered predictions, aiding in the accurate identification of the caterpillar. 
   %
   }
  \label{fig:biases}  
  \label{fig:moti}
  \vspace{-21pt}
\end{figure*}

A task-generic prompt is both coarse and potentially ambiguous, can
result in poor segmentation when directly applied. 
To address this problem, existing methods
\cite{hu2023relax,liu2023grounding} utilize the prior knowledge
embedded in Multimodal Large Language Models (MLLMs) to infer more
detailed, instance-specific prompts, such as bounding boxes or
keywords, to guide the segmentation. 
However, these MLLMs often generate hallucinations due to object
co-occurrence priors \cite{zhou2023analyzing, zhang2023siren},
mistakenly predicting non-existent elements based on the environment
as instance-specific prompts (Fig.1(a)). 
This can mislead segmentation and degrade model performance. 
While it is common to consider MLLM's hallucinations as detrimental
and should be eradicated \cite{tang2023generalization}, this phenomenon actually demonstrates a MLLM's significant capacity for contextual inference based on prior training. 
We want to explore MLLM hallucinations as a valuable untapped knowledge resource for scene understanding, critical in complex segmentation scenarios. 
In practice, when task-related objects are not prominently visible, hallucinations can fill in missing information with plausible predictions based on learned patterns of association. 
Moreover, they can also extend beyond these familiar patterns, exploring and identifying new relationships within the data that were not explicitly taught during training. This dual ability to replicate and innovate makes hallucinations a valuable asset for enhancing model performance in complex or new situations.
This predictive reasoning capacity not only fills perceptual gaps but also enriches the model's understanding, as hallucinations utilize prior knowledge to replicate and discover new patterns, enhancing insight into the target domain (see Fig.1(b)). 
Despite the potential benefits, using hallucinations to extract useful information from images to aid task remains unexplored.

\begin{figure*}[ht]
  \centering
  \includegraphics[width=14.2cm]{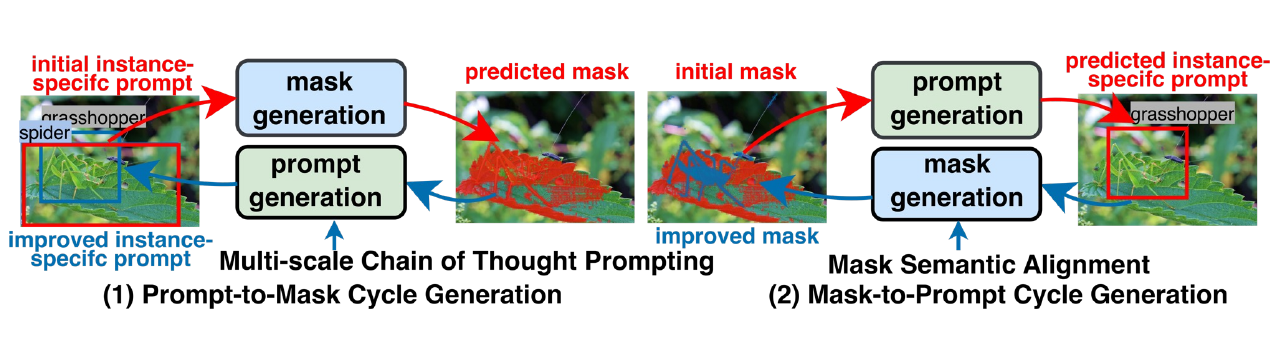}
   \vskip -0.5cm
   \caption{
    An overview of ProMac: Masks created iteratively by the mask generator guide the prompt generator to jointly improve 
    instance-specific prompts and visual masking in segmentation. 
   } 
  \label{fig:frame}
  \vspace{-10pt}
\end{figure*}

In this work, instead of direct eliminating hallucinations, we utilize them as prior knowledge to mine extended task-related information from a given test image, performing scene understanding on the
image before segmentation, then systematically reduce irrelevant hallucinations iteratively by visual masking verification, optimizing jointly instance-specific prompts and masks.
To this end, we introduce an iterative, training-free Prompt-Mask Cycle Generation method (ProMaC) that refines segmentation through cyclic interactions between a prompt and mask generator (see Fig. \ref{fig:frame}).
The prompt generator uses a multi-scale chain-of-thought prompting mechanism, which utilizes hallucinations to hypothesize and visual masking to verify, thereby creating more accurate instance-specific prompts.
We trigger the hallucinatory tendencies of MLLMs, the process starts by dividing the image into patches at different scales and positions. Such partial visibilities of objects facilitate MLLMs to hypothesize potential object semantic labels and visual locations based on its prior knowledge. 
For validating the correctness of these hypotheses, we formulate a visual contrastive reasoning mechanism to generate contrastive images that contain only the background without any potential task-related objects. This helps identify all possible co-occurrence hallucinations caused by the background. 
By comparing these contrastive images with the original images, the MLLM effectively distinguishes between accurate hypotheses and those influenced by misleading prior knowledge, leading to more reliable prompts.
Given the current promptable segmentation models' strength at mask prediction but struggle with label prediction, the mask generator uses mask semantic alignment to ensure that the produced masks align with the task semantics.
These aligned masks not only serve as outputs but also guide the prompt generator in subsequent cycles, enhancing both prompt and mask quality continuously.
%
\textbf{Our contributions are three-folds:}

1). We introduce a training-free Prompt-Mask Cycle Generation (ProMaC) to perform two tasks: Explore MLLM hallucinations as prior knowledge to enhance contextual scene understanding on each test image; systematically reduce irrelevant hallucinations to verify iteratively
and optimize jointly both generated prompts and visual masking in object segmentation. 

2).  We formulate an iterative optimization method including a prompt generator and a mask generator.
To improve prompt relevance, the prompt generator utilizes a multi-scale chain of thought approach. It first leverages
hallucinations to expand task-related plausible prompts, then applies visual contrastive reasoning to validate and reduce irrelevant prompts.
ProMaC's mask generator overcomes SAM's shortcomings in label prediction by creating masks that align semantically with generated prompts.

3). Comprehensive comparative evaluations on 5 different segmentation tasks with 12 diverse datasets against 22 existing models demonstrate the effectiveness of ProMaC. 

\section{Related Works}
\noindent\textbf{Promptable Segmentation} refers to object segmentation with active interactions from user inputs. 
Interaction methods vary from points, boxes, to scribbles. 
SAM~\cite{kirillov2023segment}, AV-SAM\cite{mo2023av}, GroundingSAM~\cite{liu2023grounding} and SEEM~\cite{zou2023segment}
accept video, audio, and multimodal inputs. However, they often rely on manual prompts, which can be unclear and subjective. Even with these prompts, they typically excel only in specific tasks.  
To address this issue, GenSAM~\cite{hu2023relax} introduces a manual-free promptable segmentation setting, where only one task-generic prompt is provided. This prompt can be applied to all images within the task for instance-specific segmentation without any additional manual prompting.
GenSAM primarily utilizes MLLM to infer the names of task-related objects in the images and uses them as instance-specific prompts for
SAM to guide segmentation. However, GenSAM lacks spatial information about objects and may lead to inaccurate prompt predictions in complex
scenes. 

\noindent\textbf{Hallucinations in MLLMs} refers to models generate
content that does not exist in the input data
\cite{zhang2023siren}. This issue often arises from the models leveraging extensive prior training rather than just the immediate
input, leading to false predictions on fine-grained details. There are some efforts to mitigate this problem, including refining training processes \cite{wang2022self1,parikh2020totto} and improving model architectures \cite{cai2021neural}. Other efforts focus on aligning model outputs more closely with actual data, employing feedback mechanisms
for real-time adjustments \cite{wan2024contrastive}. While current
works focus on eliminating hallucinations to enhance performance
\cite{leng2023mitigating,zhang2024debiasing}, our work explores how to utilize hallucinations to expand and reason plausible context and
validate them iteratively to remove irrelevant generalizations. 

\noindent\textbf{Visual Marking for MLLMs} has been explored in recent research to prompt MLLMs through manipulation of visual inputs: (i) adding learnable soft tokens to visual inputs for efficient parameter tuning \cite{bahng2022exploring, khattak2023maple}, (ii) using image sequences as demonstrations of a new task \cite{bai2023sequential, chen2023amodal}, and (iii) overlaying visual markers like masks, boxes, and circles onto visual inputs to ground regions \cite{zellers2021merlot, wan2024contrastive}.
Our work falls into the third category, employing visual guidance for reasoning.
Yang et al. \cite{yang2023set} propose set-of-mark (SoM) prompts, where images are segmented and numbered regions to improve GPT-4V \cite{openai2024gpt4v} visual grounding. 
However, as detailed in Tab.\ref{table:SR}, we confirm previous findings \cite{cai2023making} that this visual marker approach struggles with
open-source MLLMs like LLaVA.  
Instead of proprietary models \cite{wan2024contrastive} or fine-tuning \cite{cai2023making, chen2021weak, hu2020discriminative}, our training-free ProMaC uses inpainting task-related regions and contrasting model output distributions to prompt MLLMs. 

\begin{figure*}[ht]
   \centering \includegraphics[width=14cm]{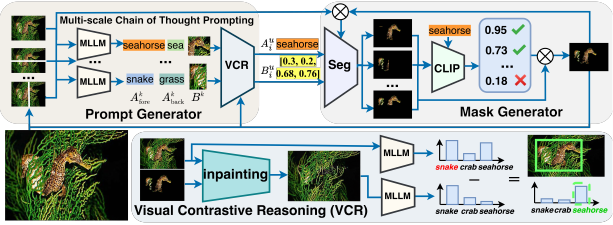}
   \caption{ProMaC consists of a prompt generator and a mask generator for cyclical optimization. The prompt generator employs multi-scale chain-of-thought prompting. It initially use hallucinations for exploring task-related information within image patches. It identifies task-relevant objects and their backgrounds ($A^k_{\text{fore}}$, $A^k_{\text{back}}$) along with their locations ($B^k$). Subsequently, it uses visual contrastive reasoning to refine and finalize instance-specific prompts ($A^u_i$, $B^u_i$) by eliminating hallucinations. The mask generator then processes these prompts into the segmentation model ("Seg"), producing a mask aligned with task semantics. This mask further guides the visual contrastive reasoning process, which leverages an inpainting model to eliminate masked regions, creating contrastive images. These images enable the prompt generator to further refine its prompts, enhancing segmentation accuracy.
   }\label{fig:framework}
\end{figure*}

\section{Methodology}
We introduce ProMaC, a cycle-generation method for segmenting unknown
multiple classes of objects training-free with only a single task-generic prompt.
Specifically, given an image \(X \in \mathbb{R}^{H \times W \times
  3}\) from a test set, ProMaC employs a task-generic prompt \(P_g\)
across datasets in the same task to produce a final segmentation mask
\(M \in \mathbb{R}^{H \times W}\), thereby removing the need for
individual supervision for each image. 
The prompt generator leverages prior knowledge gained to reason and deduce instance-specific prompts, which then guide a mask generator to create masks aligned with task
semantics. These masks act both as the current segmentation outcome
and as visual markers for generating subsequent prompts. 
Training-free ProMaC relies solely on test-time adaptation. 
%

\subsection{Prompt Generator}
\label{sec:Instance-specific Prompt Generator}
Prompt generator employs MLLMs to generate instance-specific prompts based on image content and prior knowledge. It transforms the general prompt $P_g$, into an instance-specific prompt for each individual instance, providing more detailed descriptions of task-relevant objects.
MLLM with parameters $\theta$ receives an image $X$ and query $P$ as inputs. 
$X$ provides contextual visual information to assist the model in generating a relevant response $y$ to the query $P$. The response $y$ is sampled auto-regressively from the probability distribution conditioned on $P$ and $X$ as follows:
\begin{equation}
\label{eq:mllm}
    y_t \sim p_{\theta}(y_t \mid X, P, y_{<t}) \propto \exp (\text{logit}_{\theta}(y_t \mid X, P, y_{<t}))
\end{equation}
where $y_t$ denotes the token at time step $t$, and  $y < t$ represents the sequence of generated tokens up to the time step $(t-1)$.
In practice, predicted task-relevant objects can often blend into the
background due to texture, color, size, or position,
leading to inaccuracies in instance-specific prompts.  
To address this problem, we explore MLLM hallucinations as contextual
prior knowledge from pretraining, rather than eliminate them.
These hallucinations are particularly useful when direct visual cues are absent or ambiguous, helping the model fill in information gaps and hypothesize potential task-related elements within the image that are not prominent. 
By revealing these often-overlooked subtle associations,
hallucinations provide a more comprehensive scene understanding of the image content.
This deeper contextual understanding provide a reasoning context for
generating more accurate and relevant instance-specific prompts
candidates.Thus, using hallucinations to uncover task-related
knowledge helps overcome challenges from visual ambiguities and object camouflage in complex scenes.
To this end, we propose a multi-scale chain-of-thought prompting strategy that stimulates hallucinations to leverage prior knowledge, fully extracts task-relevant information, and then uses this information to enhance the precision of the generated instance-specific prompts.
%
\subsubsection{Multi-scale Chain of Thought Prompting}
\label{sec:multi-scale chain of thought prompting}
Multi-scale Chain of Thought Prompting consists of two processes:
Gathering candidate knowledge and generating accurate
instance-specific prompts.
To efficiently collect task-relevant information from an image, as shown in Fig. \ref{fig:framework}, we divide the input image into patches at various scales by cutting horizontally, vertically, or by  leaving it whole.
These patches are then processed by the MLLM to gather preliminary instance-specific prompts.
The differing levels of task-relevant object visibility in each patch prompt the MLLM to induce hallucinations.
These hallucinations utilize prior knowledge to explore connections between the image data and the associated task, aiding in the detection of potential bounding boxes and object names. 
The process is computed by:
\begin{equation}
\label{eq: naive_bbox_object}
    B^{k} =\operatorname{MLLM}\left(X^k, C^k, P_{B}\right), \quad \quad
    A^{k}_\text {fore}, A^{k}_\text {back}=\operatorname{MLLM}\left(X^k, C^k, P_{A}\right),
\end{equation}
where $C^k$ is the caption generated by MLLM for the $k-$th image patch $X^k$. 
$P_g$ is task-generic prompt.
For bounding box prediction, the prompt $P_B$, which instructs ``{\em
  This image is from the $P_g$ detection task, output the bounding box
  of the $P_g$.}''. This guides the MLLM to predict the bounding box
$B^k$ of the task-related objects within the patch.  
For predicting name, the prompt $P_A$, stating ``{\em Output the name
  of the $P_g$ and its environment in one word.}'' is used, guiding
the MLLM to predict the names of the task-related objects
$A_{\text{fore}}^k$ and their backgrounds $A_{\text{back}}^k$ from
each patch. 
The preliminary data, including object names $A^k_{\text{fore}}$ and bounding boxes $B^k$, gathered from various patches, are compiled into candidate lists $\mathcal{A}_i$ and $\mathcal{B}_i$. Here, $i$ denotes the iteration in the iterative learning cycle.
In this process, the hallucinations employed are essentially based on object co-occurrence priors, where objects commonly associated with
background elements during pre-training are predicted to be task-relevant, even if they are not present in the current image. 
This prior knowledge is useful during the knowledge collection stage as it uncovers implicit relationships and details in the
image.
However, it can also reduce accuracy of the later fine-grained instance-specific prompts generation.
Therefore, it is crucial to control these hallucinations in the latter stage to prevent incorrect predictions.

\noindent\textbf{Visual Contrastive Reasoning.}
\label{sec:visual contrastive reasoning}
To mitigate hallucinations caused by object co-occurrence priors, recent research highlights particularly relevant regions of an image to direct MLLMs focus toward task-related elements, thereby minimizing background interference and enhancing model accuracy \cite{yang2023set, wan2024contrastive}. 
To achieve this, visual markers are employed to steer MLLM attention on task-relevant visual regions, thereby reducing hallucinations. 
While closed-source MLLMs like GPT-4V \cite{openai2024gpt4v} can
interpret these markers effectively, they are costly and large.  
In contrast,  models like LLaVA \cite{liu2023visual} are open-source, but cannot process visual markers such as points or bounding boxes, and employing these markers might disrupt the original pixel data, degrading performance on LLaVA (see Tab. \ref{table:SR}).
Moreover, accurate pixel-level visual markers are unavailable in our setting. 
To solve this problem, we aim to enable LLaVA to focus on task-related regions without altering the original pixel data, thereby effectively
minimizing hallucinations and enhancing the precision of instance-specific prompts. 

Despite the absence of instance-level annotations, promptable segmentation models produce masks with detailed textures, which provide rich positional and textural information about interested regions. 
We use these masks as visual markers to guide a MLLM to focus on task-related areas during the generation of instance-specific prompts. 
Inspired by classifier-free guidance \cite{ho2022classifier, sanchez2023stay}, we introduce visual contrastive reasoning (VCR), a training-free visual marking method to help MLLM focus on specific regions, reducing hallucinations. 
The relevance of a region is assessed by observing MLLM output changes when key areas are excluded. 
It guides the MLLM to focus on areas with notable changes (bottom of Fig.\ref{fig:framework}). 
Based on Eq.~(\ref{eq:mllm}), we derive a probability distribution by comparing original image $X$ with a modified image, $X' = \text{process}(X, \text{IM})$, where region $\text{IM}$ is excluded.
\begin{align}
y_t &\propto p_\theta(y_t \mid X, P, y_{<t}) \left( \frac{p_\theta(y_t \mid X, P, y_{<t})}{p_\theta(y_t \mid \text{process}(X, \text{IM}), P, y_{<t})} \right)^{\alpha} \notag \\
&\sim \text{softmax}[(1 + \alpha) \cdot \text{logit}_\theta(y_t \mid X, P, y_{<t}) - \alpha \cdot \text{logit}_\theta(y_t \mid \text{process}(X, \text{IM}), P, y_{<t})],
\end{align}
where $\alpha$ is the level of focus on region $\text{IM}$. A higher $\alpha$ increases emphasis on that region. Following \cite{wan2024contrastive}, we set $\alpha=1$ in all tasks.
%
%
It preserves the integrity of the original image pixels $X$, while constructing contrastive samples $X'$ that encourage the model to focus on task-related regions.
Ideally, $X'$ should exclude task-related objects while maintaining a uniform appearance and overall context with the original image.
But directly marking $X'$ disrupts its pixels, making contrastive sample generation challenging.  

\noindent\textbf{Contrastive Sample Generation.}
\label{sec:Contrastive Sample Generation}
To address it, we employ inpainting, where the mask $M_{i-1}$ obtained from the previous iteration segmentation is treated as the inpainting mask $\text{IM}_i$ to guide the creation of $X'$. 
We use a negative prompt $P_n$: "{\em $A_{i}^\text {fore}$, is a $P_g$}", to ensure that the inpainted $X'$ does not contain potentially task-related objects $A_{i}^\text {fore}$. 
Additionally, we use a positive prompt $P_p$: "{\em${A_{i}^\text {back}}$, high quality, detailed, blended to the original image.}",  to ensure consistency between the generated portion and the surrounding background ${A_{i}^\text {back}}$.
The corresponding inpainting is defined as:
\begin{equation}
     X^{\prime} = F_{in}(X, \text{IM}_i, P_p, P_n),
\end{equation}
where $F_{in}$ represents the inpainting module, and we choose Stable Diffusion to perform this operation. 
This method ensures the generated $X'$ excludes task-related objects without disrupting the pixel continuity.
In the first iteration, since $\text{IM}_{i}$ does not yet exist, we use bounding box predictions from various patches $\mathcal{B}_i$ as an alternative.
As $X'$ contains only the background, comparing it with $X$ eliminates co-occurrence hallucination caused by the background and highlights differences in task-related regions, subtly guiding the model to focus on these areas.
Finally, we use visual contrastive reasoning to identify accurate instance-specific prompt to guide segmentation as follows,
\begin{equation}
\label{eq:sum caculation}
 B_i^u =\operatorname{VCR}\left(X, X', C, P_{B}\right), \quad \quad
    A^u_{i}=\operatorname{VCR}\left(X, X', C, P_{A}\right),
\end{equation}
where $\operatorname{VCR}$ represents our visual contrastive reasoning, and $C$ is the caption of the image. The collected knowledge, $\mathcal{A}i$ and $\mathcal{B}i$, is integrated into the prompt $P_{A}$ and $P_{B}$.  This process aids in identify the ultimate instance-specific names $A^u_{i}$ and bounding boxes $B^u_{i}$ of the objects.
\subsection{Mask Generator}
\label{sec: Mask Semantic Adaptor}
Until now, we described how to use SAM-generated masks as a visual marker to guide the model to focus on task-relevant areas for generating accurate instance-specific prompts.
But this method relies on an assumption that the mask accurately delineates task-related regions.
However, SAM is trained on large-scale prompt-mask pairs without category labels, it excels at identifying masks based on image textures but lacks label prediction capabilities.
Consequently, the SAM-generated mask may not always align with task semantics, yet such alignment is crucial for our method.

\subsubsection{Mask Semantic Alignment}  
We need to utilize texture generalization capabilities of SAM to describe possible task-related objects within the prompt-targeted areas, while also ensuring that the generated masks align with the task semantics. 
To achieve this, we divide the input image into patches of varying scales using horizontal, vertical, and uncut divisions as outlined in the last section.
these processed patches are then reintegrated onto the original image with surrounding areas blacked out, and fed into SAM to ensure it focuses exclusively on the patch.
Finally, masks generated from different patches are aggregated based on their relevance to task semantics, providing an accurate representation of task-related objects.
The masks for each patch is generated as follows,
\begin{equation}
    m_i^k = \text{SAM}(\text{Spatial CLIP}(A_i^u, X_i), B_i^u, X_i^k),
\end{equation}
where mask $m_i^k$ is obtained by inputting the corresponding image patch $X_i^k$ and associated prompts into SAM during the $i-$th iteration. Following \cite{hu2023relax}, Spatial CLIP maps the text prompt $A_i^u$ to regions in the image $X_i$ that correspond to the content of the prompt.
The processed images, along with the generated instance-specific text prompts $A_i^u$, are then input into CLIP to assess semantic similarity.
\begin{equation}
    s(m_i^k) = \text{CLIP}(m_i^k \odot X_i, A_i^u),
\end{equation}
 the operation $\odot$ results in retaining only those parts of $X_i$ that are covered by the predicted mask.
$s(m_i^k)$ represents the similarity between masked image and $A_i^u$, calculated using CLIP. The similarity scores obtained from different patches are denoted as $S_i = [s(m_i^1), s(m_i^2), \ldots, s(m_i^k)]$. After normalizing the elements within $S_i$, the closer the normalized $s(m_i^k)$ is to 1, the more semantically aligned $m_i^k$ is with the instance-specific text prompt $A_i^u$. Finally, we compute the weighted sum of the normalized $s(m_i^k)$ and $m_i^k$ as follows.
\begin{equation}
    M_i = \sum_{k=1}^{K}(s(m_i^k) * m_i^k),
\end{equation}
$M_i$ is the output mask of the $i-$th iteration of $X$. The generated $M_i$ leverages SAM's mask prediction capabilities to create highly detailed masks. Simultaneously, through the mask semantic alignment process, it ensures that the output mask aligns with the task's semantics, thereby overcoming the limitation of SAM's mask prediction lacking semantic understanding.
The mask is applied to the original image as a weight, to generate the next iteration image $X_i$ for segmentation. This excludes irrelevant regions to reduce interference during segmentation.
\begin{align}
    X_{i+1}=w \cdot\left(X_i \odot M_i\right)+(1-w) \cdot X_i,
\end{align}
where $w$ is a hyperparameter, which we have assigned a value of 0.3.

\subsection{Mask Prompt Cycle Generation}
The mask generated from the last iteration will guide the prompt
generator in the next iteration to focus on potential task-related
regions, eliminating the erroneous effects of irrelevant
hallucinations and generating more accurate instance-specific prompts.  
These prompts, in turn, help the mask generator produce better masks.
Through iterative prompt generation and mask generation jointly, we
yield both better instance-specific prompts and visual masks. 
Finally, the masks from different iterations are averaged, and the mask closest to the mean is considered the final output.
\begin{equation}
\mathrm{i}^* = \arg\min_{i}\left(\left| M_i - \frac{\sum_{i}{ (M_{1}, \ldots, M_{\mathbf{I}}})}{i_{\text{result}}} \right|\right).
\end{equation}
Here, $\mathbf{I}$ is the number of adaptation epoches and $M_{\mathrm{i}^*}$ is the corresponding final mask for image $X$.

\begin{table*} 
    \centering
    \setlength{\tabcolsep}{4pt}
    \caption{Results on Camouflaged Object Detection (COD) under different settings. Best are in \textbf{bold}.}
\label{tab:results}
 \renewcommand{\arraystretch}{0.9}
 \resizebox{1.0\textwidth}{!}{
\begin{tabular}{c|c|cccc|cccc|cccc}
\cline{1-14}
{\multirow{3}{*}{Methods}} & \multicolumn{13}{c}{Camouflaged Object Detection}\\\cline{2-14}
& {\multirow{2}{*}{Venue}} &\multicolumn{4}{c|}{
CHAMELEON~\cite{skurowski2018animal}}&\multicolumn{4}{c|}{
CAMO~\cite{le2019anabranch}}&\multicolumn{4}{c}{
COD10K~\cite{fan2021concealed}}\\\cline{3-14}
& & \small{$M\downarrow$} & \small{$F_{\beta}\uparrow$} & \small{$E_{\phi}\uparrow$} &  \small{$S_{\alpha}\uparrow$}& \small{$M\downarrow$} & \small{$F_{\beta}\uparrow$} & \small{$E_{\phi}\uparrow$} &  \small{$S_{\alpha}\uparrow$} & \small{$M\downarrow$} & \small{$F_{\beta}\uparrow$} & \small{$E_{\phi}\uparrow$} &  \small{$S_{\alpha}\uparrow$} \\
\hline
\multicolumn{14}{c}{Scribble Supervision Setting} \\
\hline
WSSA\cite{zhang2020weakly} & \small{CVPR20} & 0.067 & 0.692 & 0.860 & 0.782 & 0.118 & 0.615 & 0.786& 0.696 & 0.071 & 0.536 & 0.770 & 0.684 \\
SCWS\cite{yu2021structure} & \small{AAAI21} & 0.053 & 0.758 & 0.881 & 0.792 & 0.102 & 0.658 & 0.795 & 0.713 & 0.055 & 0.602 & 0.805 & 0.710\\
TEL\cite{zhang2020weakly} & \small{CVPR22} & 0.073 & 0.708 & 0.827 & 0.785 & 0.104 & 0.681 & 0.797 & 0.717 & 0.057 & 0.633 & 0.826 & 0.724 \\
SCOD\cite{he2023weakly} & \small{AAAI23} & \textbf{0.046} & \textbf{0.791} & \textbf{0.897} & 0.818 & \textbf{0.092} & 0.709 & 0.815 & 0.735 & 0.049 & 0.637 & 0.832 & 0.733 \\
SAM-S\cite{kirillov2023segment} & \small{ICCV23} & 0.076 & 0.729 & 0.820 & 0.650 & 0.105 & 0.682 & 0.774 &0.731 & 0.046 & 0.695 & 0.828 & 0.772 \\
WS-SAM\cite{he2023weakly1}& \small{NeurlPS23} & \textbf{0.046} & 0.777 & \textbf{0.897} & \textbf{0.824} & \textbf{0.092} & \textbf{0.742} & \textbf{0.818} & \textbf{0.759} & \textbf{0.038} & \textbf{0.719} & \textbf{0.878} & \textbf{0.803}\\
\hline
\multicolumn{14}{c}{Point Supervision Setting} \\
\cline{1-14}
WSSA\cite{zhang2020weakly} & \small{CVPR20} & 0.105 & 0.660 & 0.712 & 0.711 & 0.148 & 0.607 & 0.652 & 0.649 & 0.087 & 0.509 & 0.733 & 0.642 \\
SCWS\cite{yu2021structure} & \small{AAAI21} & 0.097 & 0.684 & 0.739 & 0.714 & 0.142 & 0.624 & 0.672 & 0.687 & 0.082 & 0.593 & 0.777 & 0.738\\
TEL\cite{zhang2020weakly} & \small{CVPR22} & 0.094 & 0.712 & 0.751 & 0.746 & 0.133 & 0.662 & 0.674 & 0.645 & 0.063 & 0.623 & 0.803 & 0.727 \\
SCOD\cite{he2023weakly} & \small{AAAI23} & 0.092 & 0.688 & 0.746 & 0.725 & 0.137 & 0.629 & 0.688 & 0.663 & 0.060 & 0.607 & 0.802 & 0.711 \\
SAM\cite{kirillov2023segment} & \small{ICCV23} & 0.207 & 0.595 & 0.647 & 0.635 & 0.160 & 0.597 & 0.639 & 0.643 & 0.093 & 0.673 & 0.737 & 0.730 \\
SAM-P\cite{kirillov2023segment} & \small{ICCV23} & 0.101 & 0.696 & 0.745 & 0.697 & 0.123 & 0.649 & 0.693 & 0.677 & 0.069 & 0.694 & 0.796 & 0.765\\
WS-SAM\cite{he2023weakly1} & \small{NeurlPS23} & \textbf{0.056} & \textbf{0.767} & \textbf{0.868} & \textbf{0.805} & \textbf{0.102} & \textbf{0.703} & \textbf{0.757} & \textbf{0.718} & \textbf{0.039} & \textbf{0.698} & \textbf{0.856} & \textbf{0.790}\\\hline
\multicolumn{14}{c}{Task-Generic Prompt Setting} \\ 
\cline{1-14}

{\small CLIP\_Surgey+SAM} & \small{Arxiv23} & 0.147 & 0.606 & 0.741 & 0.689 & 0.189 & 0.520 & 0.692 & 0.612 & 0.173 & 0.488 & 0.698 & 0.629	  \\
{\small GPT4V+SAM} \cite{openai2024gpt4v,kirillov2023segment} & \small{Arxiv23} &  0.180 & 0.557 & 0.710 & 0.637& 0.206& 0.466& 0.666 & 0.573 & 0.187 & 0.448 &0.672 & 0.601\\
{\small LLaVA1.5+SAM} \cite{liu2023visual,kirillov2023segment} & \small{NeurlPS23} & 0.168& 0.561 & 0.718 & 0.666& 0.314 & 0.401 & 0.585 & 0.501 & 0.170 & 0.530 & 0.728 & 0.662\\
X-Decoder~\cite{zou2023generalized}  & {\small{CVPR23}} &0.124& 0.654 & 0.748 & 0.716 & 0.104& 0.628 & 0.745 & 0.709 & 0.171 & 0.556 & 0.705 & 0.652\\
SEEM~\cite{zou2023segment} & {\small{NeurIPS23}} & 0.094&  0.011 & 0.307 & 0.454 & 0.192 & 0.023 & 0.315 & 0.404 & 0.143 & 0.001 & 0.280 & 0.425\\
GroundingSAM~\cite{kirillov2023segment, liu2023grounding} & {\small{ICCV23}} & 0.122 & 0.662 & 0.776 & 0.744 & 0.157 & 0.656 & 0.753 & 0.707 & 0.085 & 0.670 & 0.813& 0.764\\
GenSAM~\cite{hu2023relax} & \small{AAAI24} & 0.073 & 0.696 & 0.806 & 0.774 & 0.106 & 0.669 & 0.798 & 0.729 & 0.058 & 0.695 & 0.843 & 0.783\\
\rowcolor{purple!10}ProMaC & Ours & \textbf{0.044} & \textbf{0.790} & \textbf{0.899} & \textbf{0.833} & \textbf{0.090}	& \textbf{0.725} & \textbf{0.846} & \textbf{0.767} & \textbf{0.042} & \textbf{0.716} & \textbf{0.876} & \textbf{0.805}\\
\hline
\end{tabular}
}
\end{table*}

\begin{table*}[ht]
\setlength{\tabcolsep}{5pt}
    \caption{Results for Medical Image Segmentation (MIS) under task-generic prompt setting.}
    \label{tab:results_2m}
    \centering 
     \renewcommand{\arraystretch}{0.9}
  \resizebox{1.0\textwidth}{!}{
\begin{tabular}{c|c|cccc|cccc|cccc}
\hline
\multicolumn{1}{c|}{\multirow{3}{*}{Methods}} &{\multirow{3}{*}{Venue}} & \multicolumn{8}{c}{Polyp Image   Segmentation} &\multicolumn{4}{|c}{Skin Lesion Segmentation}  \\ \cline{3-14} &  &\multicolumn{4}{c|}{CVC-ColonDB~\cite{tajbakhsh2015automated}} & \multicolumn{4}{c|}{Kvasir~\cite{jha2020kvasir}}      & \multicolumn{4}{c}{ISIC~\cite{codella2019skin}}       \\ \cline{3-14}
& &\small{$M\downarrow$} & \small{$F_{\beta}\uparrow$} & \small{$E_{\phi}\uparrow$} &  \small{$S_{\alpha}\uparrow$}
&\small{$M\downarrow$} & \small{$F_{\beta}\uparrow$} & \small{$E_{\phi}\uparrow$} &  \small{$S_{\alpha}\uparrow$}
&\small{$M\downarrow$} & \small{$F_{\beta}\uparrow$} & \small{$E_{\phi}\uparrow$} &  \small{$S_{\alpha}\uparrow$}\\\hline
{\small GPT4V+SAM} \cite{openai2024gpt4v,kirillov2023segment}& {\small{Arxiv23}}& 0.578 & 0.051 &    0.246 & 0.242& 0.614 & 0.128 & 0.236 & 0.253 & 0.514 & 0.387 & 0.366 & 0.334 \\
{\small LLaVA1.5+SAM} \cite{liu2023visual,kirillov2023segment} & {\small{NeruIPS23}}& 0.491 & 0.194 & 0.355 & 0.357 & 0.479 & 0.293 & 0.400 & 0.403 & 0.369	& 0.473	& 0.497	& 0.477
\\
X-Decoder~\cite{zou2023generalized} &  {\small{CVPR23}}& 0.462 & 0.095 & 0.327 & 0.331 & 0.449 & 0.202 & 0.371 & 0.384 & 0.338 & 0.315 & 0.127 & 0.407  \\
SEEM~\cite{zou2023segment} & {\small{NeruIPS23}}& 0.570 & 0.085 & 0.280 & 0.284 & 0.520 & 0.215 & 0.339 & 0.367 & 0.362 & 0.250 & 0.002 & 0.280 \\
GroundingSAM~\cite{kirillov2023segment, liu2023grounding} & {\small{ICCV23}}&    0.711    &   0.071    &   0.195    &    0.206   &   0.387    &  0.353    &    0.521   &    0.468  & 0.301 & 0.348 & 0.247 & 0.533\\ 
GenSAM~\cite{hu2023relax}& {\small{AAAI24}} & 0.244 & 0.059	& 0.494	&0.379 & 0.172 &	0.210 &	0.619 &	0.487 & 0.171 & 0.699 &	0.744 & 0.678 \\
\rowcolor{purple!10}ProMaC & {\small{Ours}}& \textbf{0.176}	 &  \textbf{0.243} & \textbf{0.583} &\textbf{0.530} & \textbf{0.166}	& \textbf{0.394}	 & \textbf{0.726} & \textbf{0.573} & \textbf{0.168} & \textbf{0.717} & \textbf{0.755} & \textbf{0.689}\\ 
\hline
\end{tabular}}
\end{table*}

\begin{table*}[ht]
    \setlength{\tabcolsep}{1.5pt}
    \centering
    \renewcommand{\arraystretch}{1.1}
    \caption{Result on Transparent Object Segmentation and Open-Vocabulary Segmentation Tasks.
    }
    \label{tab:CombinedResults}
   {
    \begin{tabular}{@{\hspace{0pt}}c@{\hspace{50pt}}c}  
    { (a) Transparent Object Segmentation. \label{table:explainability}}&
     { (b) Open-vocabulary Segmentation.\label{table:open-vocabulary}  }
\end{tabular}}

    \resizebox{1.0\textwidth}{!}{
    \begin{tabular}{c@{\hspace{5pt}}c} 
    \begin{tabular}{c|cccc|cccc}
\hline
\multirow{2}{*}{Methods}
& \multicolumn{4}{c|}{GSD~\cite{lin2021rich}}  & \multicolumn{4}{c}{Trans10K-hard \cite{xie2020segmenting}} \\ \cline{2-9}
& $M\downarrow$ & $F_{\beta}\uparrow$ & $E_{\phi}\uparrow$ & $S_{\alpha}\uparrow$ & $M\downarrow$ & $F_{\beta}\uparrow$ & $E_{\phi}\uparrow$ & $S_{\alpha}\uparrow$ \\
\hline
GPT4V+SAM \cite{openai2024gpt4v,kirillov2023segment} & 0.312	& 0.104 &  0.392& 0.363 & 0.288 & 0.199 & 0.607 & 0.512\\
LLaVA1.5+SAM \cite{liu2023visual,kirillov2023segment} &0.197 & 0.202 & 0.545 & 0.433 &0.272 & 0.167 & 0.621 & 0.555\\
X-Decoder~\cite{zou2023generalized} & 0.191 & 0.240 & 0.643 & 0.480 & 0.568 & \textbf{0.611} & 0.218 & 0.280\\
SEEM~\cite{zou2023segment} & 0.184 & 0.224 & 0.573 & 0.479 & 0.557 & 0.501 & 0.013 & 0.256 \\
GroundingSAM~\cite{kirillov2023segment,liu2023grounding} & 0.168 & 0.230 & 0.572 & 0.483 & 0.436 & 0.415 & 0.047 & 0.424\\
GenSAM~\cite{hu2023relax} & 0.155 & 0.394 &	0.700 &	0.559  &  0.263  & 0.489  & 0.612 & 0.536 \\
\rowcolor{purple!10}ProMaC & \textbf{0.147} & \textbf{0.409} & \textbf{0.723} & \textbf{0.569} &  \textbf{0.251}	& 0.509	& \textbf{0.654}	& \textbf{0.557} \\
\hline
\end{tabular}
&
  \begin{tabular}{c|c|cc|c|c|c}
    \hline
    \multirow{2}{*}{Methods} & \multirow{2}{*}{Venue} & \multirow{2}{*}{Seg. Anno.} & \multirow{2}{*}{Image-Text pairs} & VOC & Context & Object \\
    \cline{5-7}
    &&&& mIoU $\uparrow$ & mIoU $\uparrow$ & mIoU $\uparrow$ \\ \hline
    MaskCLIP\cite{zhou2022extract} & ECCV22 & - & - & 38.8 & 23.6 & 20.6 \\
    TCL \cite{cha2023learning} & CVPR23 & - & CC3M \cite{sharma2018conceptual}, CC12M \cite{changpinyo2021conceptual} & 51.2 & 24.3 & \textbf{30.4}\\
    GroupViT \cite{xu2022groupvit} & CVPR22 & - & CC12M \cite{changpinyo2021conceptual}, YFCC14M \cite{thomee2016yfcc100m} & 52.3 & 22.4 & -\\
    ViewCo \cite{ren2023viewco} & ICLR23 & - & CC12M \cite{changpinyo2021conceptual}, YFCC14M \cite{thomee2016yfcc100m} & 52.4 & 23.0 & 23.5\\
    SegCLIP \cite{luo2023segclip} & ICML23 & COCO \cite{lin2014microsoft} & CC \cite{sharma2018conceptual} & 52.6 & 24.7 & {26.5}\\
    OVSegmentor \cite{xu2023learning} & CVPR23 & - & CC12M \cite{changpinyo2021conceptual} & 53.8 & 20.4 & 25.1 \\ 
    \rowcolor{purple!10} ProMaC & Ours & - & - & \textbf{59.3} & \textbf{30.7} & 25.2\\ 
    \hline
  \end{tabular}
\end{tabular}}
\end{table*}

\begin{table}[t]
    \centering
    \scriptsize
    \hspace{-6mm}
    \begin{minipage}[t]{0.7\textwidth}
        \centering
        \renewcommand{\arraystretch}{0.7}
        \captionof{table}{Ablation Study on COD and MIS Tasks}\label{table:ablation}
        \vspace{2pt}
        \setlength{\tabcolsep}{2pt}
        \begin{tabular}{ccccc|cccc|cccc}
            \hline
            \multicolumn{5}{c|}{Method's Variants} & \multicolumn{4}{c|}{CHAMELEON~\cite{skurowski2018animal}} & \multicolumn{4}{c}{CVC-ColobNB ~\cite{tajbakhsh2015automated}} \\
            \hline
            MCoT & IVP & ITP & VCR & MSA & $M\downarrow$ & $F_{\beta}\uparrow$ & $E_{\phi}\uparrow$ & $S_{\alpha}\uparrow$ & $M\downarrow$ & $F_{\beta}\uparrow$ & $E_{\phi}\uparrow$ & $S_{\alpha}\uparrow$ \\
            \hline
            & \checkmark & \checkmark & \checkmark & \checkmark & 0.052 & 0.764 & 0.885 & 0.816 & 0.187 & 0.214 & 0.570 & 0.513 \\
            \checkmark & & \checkmark & \checkmark & \checkmark & 0.080 & 0.720 & 0.833 & 0.757 & 0.260 & 0.123 & 0.466 & 0.425 \\
            \checkmark & \checkmark & & \checkmark & \checkmark & 0.089 & 0.685 & 0.823 & 0.756 & 0.177 & 0.233 & 0.556 & 0.524 \\
            \checkmark & \checkmark & \checkmark & & \checkmark & 0.061 & 0.769 & 0.893 & 0.815 & 0.311 & 0.152 & 0.460 & 0.424 \\
            \checkmark & \checkmark & \checkmark & \checkmark & & 0.054 & 0.740 & 0.884 & 0.798 & \textbf{0.156} & 0.220 & 0.565 & 0.517 \\
            \rowcolor{purple!10}
            \checkmark & \checkmark & \checkmark & \checkmark & \checkmark & \textbf{0.044} & \textbf{0.790} & \textbf{0.899} & \textbf{0.833} & 0.176 & \textbf{0.243} & \textbf{0.583} & \textbf{0.530} \\
            \hline
        \end{tabular}
    \end{minipage}
     \hspace{-3mm}
    \begin{minipage}[t]{0.34\textwidth}
        \centering
        \scriptsize
        \captionof{table}{VCR Result on SR task}\label{table:SR}
        \vspace{2pt}
        \renewcommand{\arraystretch}{0.925}
        \setlength{\tabcolsep}{1pt}
        \begin{tabular}{lcccc}
            \hline
            Model & Indiv. & Pairs & Set of 4 \\
            \hline
            CLIP ViT-L-14 \cite{radford2021learning} & 26.1 & 1.5 & 0.0 \\
            CLIP RN50x64 \cite{radford2021learning} & 26.2 & 2.0 & 0.0 \\
            FLAVA \cite{singh2022flava} & 30.4 & 10.9 & 0.0 \\
            ViP-LLAVA-13B \cite{cai2023making} & 70.9 & 57.5 & 21.8 \\
            \hline
            LLAVA-1.5-13B \cite{liu2023improvedllava} & 73.1 & 60.6 & 28.9 \\
            \rowcolor{purple!10}+ VCR (Ours) & \textbf{75.4} & \textbf{63.6} & \textbf{36.7} \\
            \hline
        \end{tabular}
    \end{minipage}
\end{table}

\begin{table*}[tb!]
\caption{{Parameter ablation study on COD10K~\cite{fan2021concealed}. }}
  \vskip -0.1cm
\resizebox{1.0\textwidth}{!}{
\footnotesize
  \begin{tabular}
  { @{\hspace{-10pt}}
  c @{\hspace{0pt}}  c@{\hspace{0pt}}
  c@{\hspace{0pt}}}
    { (a) Number of iteration $\mathbf{I}$.\label{table:itertion}}&
     { (b) Image preprocess strategy.\label{table:scale}}&
     {(c) Visual marker strategy. \label{table:Postprocessing}}\\
    {
    \footnotesize
    \setlength{\tabcolsep}{5pt}
    \renewcommand{\arraystretch}{0.65}
        \begin{tabular}{c|cccccc} 
        \hline
        $\mathbf{I}$ & {cos$\uparrow$}& {IoU$\uparrow$}  & {$M\downarrow$} & {$F_{\beta}\uparrow$} & {$E_{\phi}\uparrow$} &  {$S_{\alpha}\uparrow$}\\ \hline
        1  & 0.864 & 0.563 &  0.080 & 0.626 & 0.818 & 0.765 \\
        2  & 0.876& 0.589 &0.050 & 0.683 & 0.859 & 0.796  \\
        3  & 0.879 & 0.593 & 0.045 & 0.702 & 0.869 & 0.802\\
        \cellcolor{purple!10}4 & \cellcolor{purple!10}\textbf{0.882} & \cellcolor{purple!10}0.601 & \cellcolor{purple!10}0.042  & \cellcolor{purple!10}0.714 & \cellcolor{purple!10} 0.875  & \cellcolor{purple!10}\textbf{0.804} \\
        5  &0.881 &\textbf{0.602} & \textbf{0.041} & 0.718 & 0.875 & \textbf{0.804}\\ 
        6 & \textbf{0.882} & 0.599 & \textbf{0.041} & \textbf{0.721} & \textbf{0.876} & 0.803 \\\hline
    \end{tabular}}&
    {\setlength{\tabcolsep}{5pt}
    \renewcommand{\arraystretch}{0.95}
        \begin{tabular}{c|cccc} 
        \hline
        Scale      &  {$M\downarrow$} & {$F_{\beta}\uparrow$} & {$E_{\phi}\uparrow$} &  {$S_{\alpha}\uparrow$} \\ \hline
        Original   & 0.075 & 0.535 & 0.750 & 0.662  \\
        Havel   & 0.069	& 0.579	& 0.775 & 0.689  \\
        Quarters & 0.087 & 0.423 & 0.673 &  0.586 \\
        \cellcolor{purple!10}Original+Havel   & \cellcolor{purple!10}\textbf{0.042} & \cellcolor{purple!10}\textbf{0.714} & \cellcolor{purple!10}\textbf{0.875} & \cellcolor{purple!10}\textbf{0.804}  \\
        Original  +Havel+Quarters  & 0.049 & 0.702 & 0.867  & 0.796   \\ \hline
    \end{tabular}} &
    {\setlength{\tabcolsep}{5pt}
    \renewcommand{\arraystretch}{1.15}
    \begin{tabular}{c|cccc} 
    \hline
    \footnotesize{strategy}      & {$M\downarrow$} & {$F_{\beta}\uparrow$} & {$E_{\phi}\uparrow$} &  {$S_{\alpha}\uparrow$} \\ \hline
    \footnotesize{None}       & 0.058 & 0.690 &  0.855 & 0.789  \\
    \footnotesize{Bbox}   & 0.065 & 0.682 & 0.836 & 0.766   \\
    \footnotesize{VCD} & 0.047 &  0.705 & 0.863 & 0.793 \\
    \rowcolor{purple!10}\footnotesize{Ours}  & \textbf{0.042} & \textbf{0.714} & \textbf{0.875}  & \textbf{0.804}                               \\\hline
    \end{tabular}} 
     
  \end{tabular}}
\end{table*}

\section{Experimental Setup}
\label{sec:experiment}
\noindent\textbf{Baseline.} To evaluate our approach across various scenarios, we first assess the performance of ProMaC on challenging segmentation tasks, including Camouflaged Object Detection (COD), Medical Image Segmentation (MIS), and Transparent Object Detection (TOD). These tasks are areas where SAM struggle \cite{ji2023segment}.
In the COD task, we compare ProMaC with weakly supervised segmentation methods \cite{kirillov2023segment,zhang2020weakly,yu2021structure,zhang2020weakly,he2023weakly,he2023weakly,hu2019multi,hu2022learning, he2023camouflaged}. 
Two supervision levels are used for comparison: scribble supervision, where main structures for the foreground and background are sketched during training, and point supervision, where separate points are provided for both foreground and background.
In the task-generic prompt setting, we introduce a challenging scenario by only providing a task description as a generic prompt for segmentation. ProMaC integrates LLaVA1.5 \cite{liu2023visual} with SAM \cite{kirillov2023segment}.
We also experiment on the MIS and PIS tasks to demonstrate the effectiveness of our method using task-generic prompts compared to previous methods.
We assess GPT4V+SAM and LLaVA1.5+SAM in this setting to demonstrate that current MLLM models cannot address it well. We also compare ProMaC against current SOTA promptable segmentation methods to showcase its effectiveness.
Next, we evaluate ProMaC on Open-Vocabulary Segmentation (OVS), and compare with leading methods \cite{li2023clip, zou2022xdecoder, zou2023segment, liu2023grounding,zhang2021domain,zhong2019transfer,li2023uncertainty,he2023strategic}.
Our Visual Contrastive Reasoning (VCR) strategy is applicable to other tasks, especially those requiring complex spatial understanding. We used the What's Up spatial reasoning dataset \cite{kamath2023s} to evaluate how well VCR guides models to focus on task-relevant regions in Spatial Reasoning (SR) task.
Our results are the average of three trials.

\noindent\textbf{Metric.} For evaluating the first three tasks, metrics Mean Absolute Error (M), adaptive F-measure (\( F_{\beta} \)) \cite{margolin2014evaluate}, mean E-measure (\( E_{\phi} \)) \cite{fan2021cognitive}, and structure measure (\( S_{\alpha} \)) \cite{fan2017structure} are used. Lower M or higher values for \( F_{\beta}, E_{\phi}, \) and \( S_{\alpha} \) reflect better performance. Mean Intersection over Union (mIoU) and accuracy measure OVS and SR performance respectively, with higher values indicating better results.

\noindent\textbf{PyTorch Implementation Details.}
For the MLLM models, we utilize LLaVA-1.5-13B for evaluation purposes. 
For CLIP, we adopt the CS-ViT-B/16 pretrained model. 
The inpainting model is stable-diffusion-2-inpainting.
The task-generic prompts for the COD task is "camouflaged animal". 
The MIS task consists of two sub-tasks: polyp image segmentation and skin lesion segmentation, each with its own task-generic prompts, "polyp" and "skin lesion" respectively.
For TOD task, prompt is fixed as "glass".
All tasks are optimized using training-free test-time adaptation, with each task iterating for four epochs, except for the polyp image segmentation task, which undergoes six epochs.
Since the second epoch, VCR also operates on different patches, ensuring that while the non-inpainted parts still gather information through hallucinations, the inpainted parts eliminate hallucinations to generate accurate candidate prompts.
The promptable segmentation methods is the ViT-H/16 version of SAM.
Our experiment is conducted on a single NVIDIA A100 GPU. More details are in appendix.
\begin{figure*}[ht]
   \centering   \includegraphics[width=14cm]{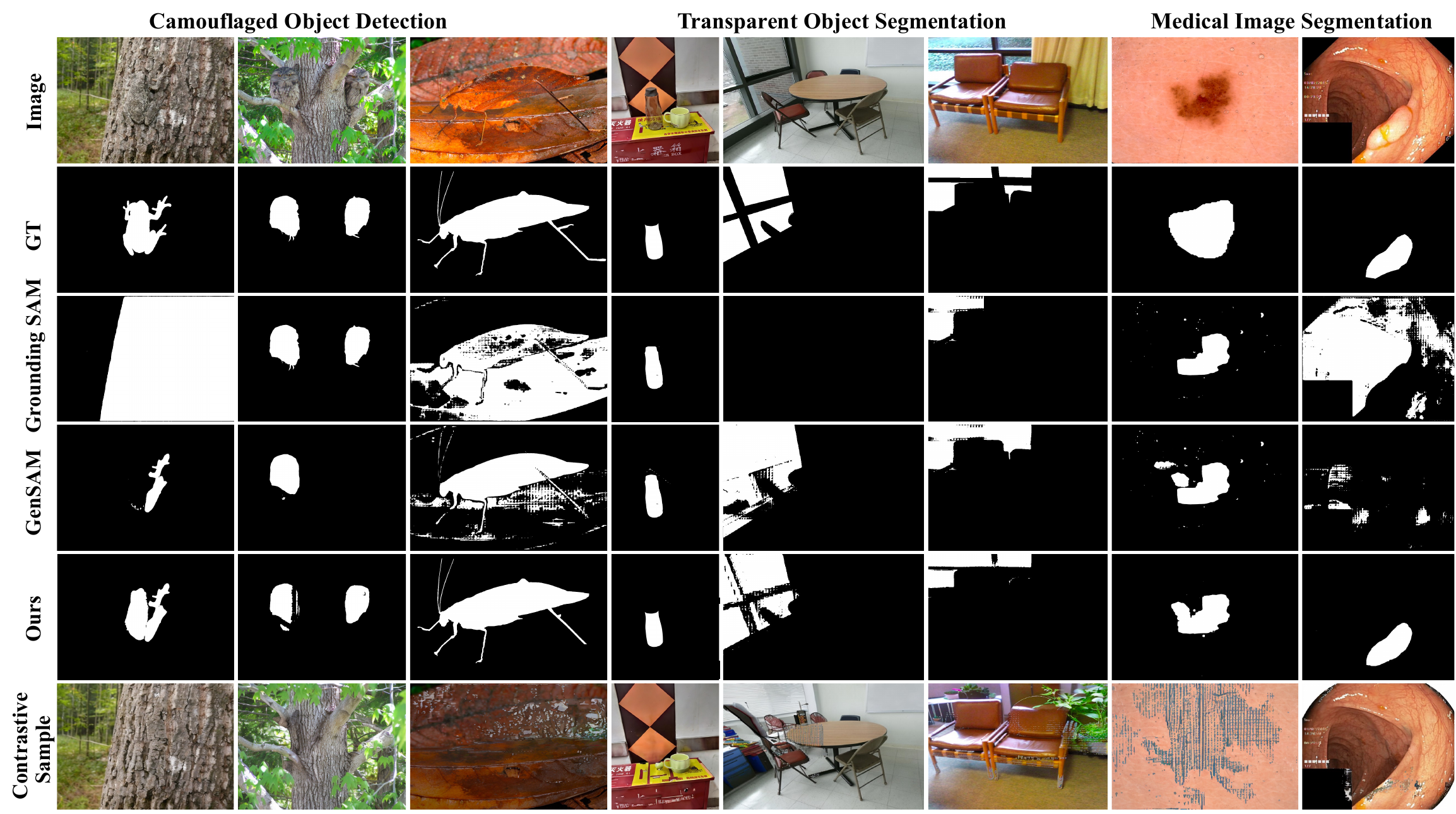}
   \caption{Visualization of various segmentation methods among various segmentation tasks.
}\label{fig:visualization}
\vspace{-6pt}
\end{figure*}

\begin{figure*}[htbp]
   \centering   
   \includegraphics[width=14cm]{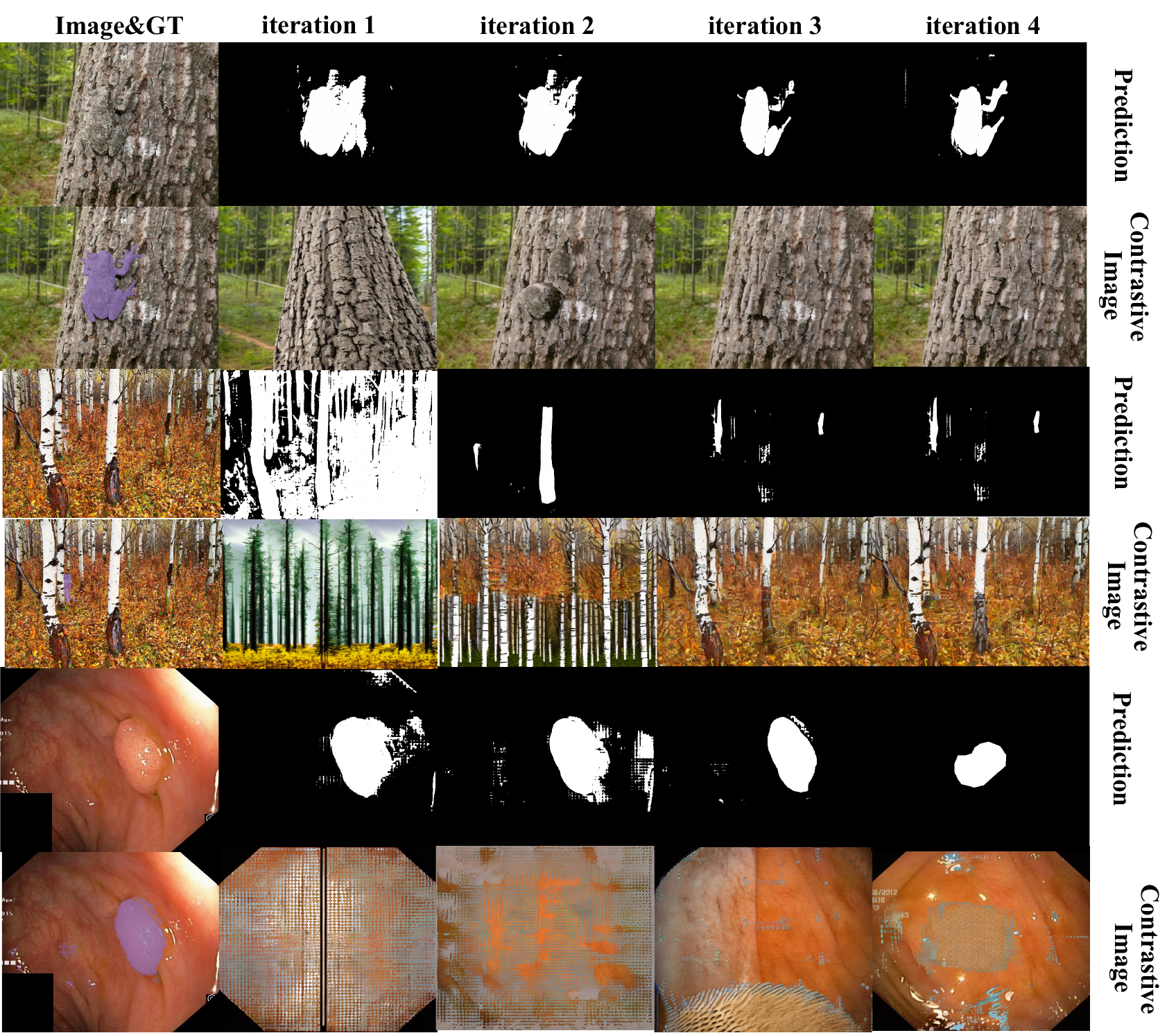}
   \caption{Visualization of the generated masks and contrastive samples over iterations.}
   \label{fig:visualization_iter}
   \vspace{-10pt}
\end{figure*}

\vspace{-15pt}
\section{Results and Analysis}
\noindent\textbf{Results on COD Task.} The COD tasks focus on finding animals that blend into their complex surroundings.
We evaluated ProMaC on three representative datasets: CHAMELEON \cite{skurowski2018animal}, CAMO \cite{le2019anabranch}, and COD10K \cite{fan2021concealed}.
As shown in Tab. \ref{tab:results}, we compared ProMaC with others that utilize varying levels of supervision. Overall, methods with scribble supervision generally perform better than those with point supervision. 
Importantly, ProMaC only uses a single generic task prompt for the
entire task and it stil outperforms all point-supervised methods. It also surpasses methods with scribble supervision on the CHAMELEON and CAMO datasets, and matches the top-performing scribble-supervised methods on COD10K. 
It demonstrates the superiority of ProMaC.

\noindent\textbf{Results on MIS and TOD Task.}
The MIS task identifies pathological tissues in medical images. We
used three datasets: ColonDB \cite{tajbakhsh2015automated} and Kvasir
\cite{jha2020kvasir} for polyp image segmentation, and ISIC
\cite{codella2019skin} for skin lesion segmentation. 
We compared our approach with others using task-generic prompt settings (see Tab. \ref{tab:results_2m}). While other models underperform in medical imaging due to limited generalization, ProMaC improves significantly over the baseline by iteratively mining task-related knowledge. 
For the TOD task, we evaluated ProMaC on the GSD \cite{lin2021rich} and Trans10K-hard \cite{xie2020segmenting} datasets (See Tab. \ref{tab:CombinedResults}(a)). Using the task-generic prompt setting, our method achieves the best results despite challenging scenarios. This demonstrates ProMaC's versatility and adaptability across complex visual tasks.

\noindent\textbf{Results on OVS and SR Task.}  
We evaluated ProMaC's effectiveness on the OVS task for multi-class segmentation based on a list of candidate classes. 
Specifically, we tested it on the validation splits of PASCAL VOC (21 classes) \cite{everingham2010pascal, everingham2012pascal}, Pascal Context (59 classes) \cite{mottaghi2014role}, and COCO-Object (80 classes) \cite{caesar2018coco}, using LLaVA to identify and confirm the presence of candidate classes. After obtaining masks, we resolved overlaps using the argmax operation based on SAM probabilities. 
Tab. \ref{tab:CombinedResults}(b) shows how ProMaC compares to other state-of-the-art OVS methods. Unlike some methods trained specifically
on these datasets (risking knowledge leaking), ProMaC is not. Yet, ProMaC still outperforms all others on PASCAL VOC and Pascal Context and is competitive on COCO-Object.
Additionally, as shown in Tab. \ref{table:SR}, we integrated our VCR into LLaVA1.5 for enhanced spatial reasoning. This integration allows LLaVA to better focus on critical areas, thereby boosting performance. 

\noindent\textbf{Module Analysis.} As shown in Tab. \ref{table:ablation}, we perform an ablation study on the COD and MIS tasks to assess the effects of different modules.
"MCoT" is multi-scale chain of thought prompting. "ITP" and "IVP" refer to using only instance-specific text prompts or visual prompts. "VCR" is visual contrastive reasoning, and "MSA" is mask semantic alignment. 
The first row shows replacing MCoT with just one original image results in reduced performance, highlighting the importance of using hallucinations to extract task-relevant information.
The second and third rows show that single modal prompts perform worse than multimodal prompts, highlighting the significance of multimodal prompting.
Removing VCR causes a significant drop in performance, indicating that visual prompts are crucial for directing LLaVA's focus on relevant areas during inference.
The comparison between the fifth and final rows emphasizes the importance of mask alignment with task semantics. The consistent positive results across tasks confirm the robustness and effectiveness of our approach.

\noindent\textbf{Parameter Analysis.} 
Tab. \ref{table:itertion}(a) examines how iterations influence performance. 
"cos" measures the cosine similarity between the predicted text prompt and the ground truth class through CLIP. 
"IoU" assesses the overlap between the predicted bounding box and the ground truth, comparing it against a rectangular outline of the mask. 
Mask predictions improve and stabilize after the fourth epoch.
Tab. \ref{table:scale}(b) investigates the effects of various image processing techniques. 
"Original" uses no modifications, "Halve" divides the image horizontally or vertically into halves, and "Quarters" divides it into four quarter-sized patches. Testing shows that combining "Original" and "Halve" yields the best results by balancing global and local information without excessive fragmentation.

\noindent\textbf{Visual Marker Strategy.} Tab. \ref{table:Postprocessing}(c) assesses the impact of different visual marker strategies. 
"None" uses no visual prompts, while "Bbox" places bounding boxes directly on the image. 
"VCD" employs previous methods that introduce Gaussian noise into comparison images for contrastive reasoning. 
Results indicate that bounding boxes decrease performance, suggesting LLaVA struggles with this type of markers. 
Although VCD methods improve performance, they distort pixel data, making them less effective than our approach. 
Our VCR generates contrastive samples that focus on task-relevant areas without altering the image, reducing hallucinations and enhancing performance.

\noindent\textbf{Visualization.} Fig. \ref{fig:visualization} and Fig.\ref{fig:visualization_iter} visually compares our ProMaC with other methods across 3 tasks and also shows the contrastive images we generated. 
GenSAM handles clear objects well but struggles with complex background. 
Although GenSAM performs well in complex backgrounds, but struggles with challenging tasks. 
ProMaC delivers solid segmentation results across different tasks, and our contrastive images remove task-related regions while maintaining semantic and pixel consistency.

\section{Conclusion}
In this work, we introduce an iterative ProMaC that uses MLLM hallucinations to guide automatic prompt generation, significantly improving segmentation without training. This iterative approach aligns masks with task semantics, enhancing model
performance. Testing on multiple benchmarks has demonstrated ProMaC's effectiveness in a wide range of complex segmentation tasks. 


{\bibliographystyle{unsrt}

\newpage
\appendix

\end{document}